\documentclass[10pt,twocolumn,letterpaper]{article}

\usepackage[pagenumbers]{iccv} % To force page numbers, e.g. for an arXiv version

\definecolor{iccvblue}{rgb}{0.21,0.49,0.74}
\usepackage[pagebackref,breaklinks,colorlinks,allcolors=iccvblue]{hyperref}
\usepackage{tikz}
\usepackage{caption}
\usepackage{graphicx}  

\def\paperID{****} % *** Enter the Paper ID here
\def\confName{ICCV}
\def\confYear{2025}

\title{Attention-Aware Multi-View Pedestrian Tracking}

\author{Reef Alturki\hspace{1cm}Adrian Hilton\hspace{1cm}Jean-Yves Guillemaut\\
Centre for Vision, Speech and Signal Processing, University of Surrey, UK.\\
{\tt\small \{r.alturki, a.hilton, j.guillemaut\}@surrey.ac.uk}
}
\begin{document}
\maketitle
\begin{abstract}
In spite of the recent advancements in multi-object tracking, occlusion poses a significant challenge. Multi-camera setups have been used to address this challenge by providing a comprehensive coverage of the scene. Recent multi-view pedestrian detection models have highlighted the potential of an early-fusion strategy, projecting feature maps of all views to a common ground plane or the Bird's Eye View (BEV), and then performing detection. This strategy has been shown to improve both detection and tracking performance. However, the perspective transformation results in significant distortion on the ground plane, affecting the robustness of the appearance features of the pedestrians. To tackle this limitation, we propose a novel model that incorporates attention mechanisms in a multi-view pedestrian tracking scenario. Our model utilizes an early-fusion strategy for detection, and a cross-attention mechanism to establish robust associations between pedestrians in different frames, while efficiently propagating pedestrian features across frames, resulting in a more robust feature representation for each pedestrian. Extensive experiments demonstrate that our model outperforms state-of-the-art models, with an IDF1 score of $96.1\%$ on Wildtrack dataset, and $85.7\%$ on MultiviewX dataset.
\end{abstract}    
\section{Introduction}
\label{sec:intro}
Tracking objects from a single camera has attracted significant attention in the research community. However, crowded environments and heavy occlusions continue to pose significant challenges in single-camera setups due to the inherent limitations in capturing comprehensive scene information. Occlusion leads to missed detections and fragmented tracks, negatively impacting the performance of detection and tracking models. Multi-camera setups offer a promising solution to tackle this challenge by providing comprehensive coverage of the scene. Such setups often utilize cameras with overlapping fields of view, allowing objects that are occluded in one camera to be visible by another.

Recent approaches in multi-view detection utilize an early-fusion strategy, projecting feature maps from all views to the ground plane or the Bird's Eye View (BEV) to create a unified representation of the scene where the detection analysis is performed \cite{ref28, ref29}. This strategy has been shown to significantly enhance detection accuracy compared to late-fusion approaches. Moreover, it has become a dominant approach in the detection step of the recent multi-view tracking models \cite{ref36, ref38, ref39}. In \cite{ref36}, the tracking process primarily relies on re-ID features extracted from the BEV feature for each detection. However, the perspective projection results in substantial distortion on the ground plane, negatively impacting the robustness of the appearance features, which are essential for establishing accurate temporal associations between pedestrians across frames. 

\iffalse
\begin{figure}[t]
    \centering
    \includegraphics[width=0.46\textwidth]{images/illu1.png} 
    \caption{Overview of the cross-attention module. It takes BEV features from two frames and extracts the features at pedestrian locations, which are used as tokens for the cross-attention processing. The module computes affinity scores between pedestrians in different frames and propagates the features across frames.  \begin{tikzpicture} \draw[scale=0.5] (0,0) circle (0.23cm); \draw[scale=0.5,domain=-0.25:0.25,samples=200] plot (\x,{0.1*-sin(10*\x r)}); \end{tikzpicture} represents positional encoding, while $\otimes$ and $\odot$ denote element-wise and matrix multiplication, respectively. }
    \label{fig:fig1}
\end{figure}
\fi

\begin{figure}[t]
    \centering
    \includegraphics[width=0.46\textwidth]{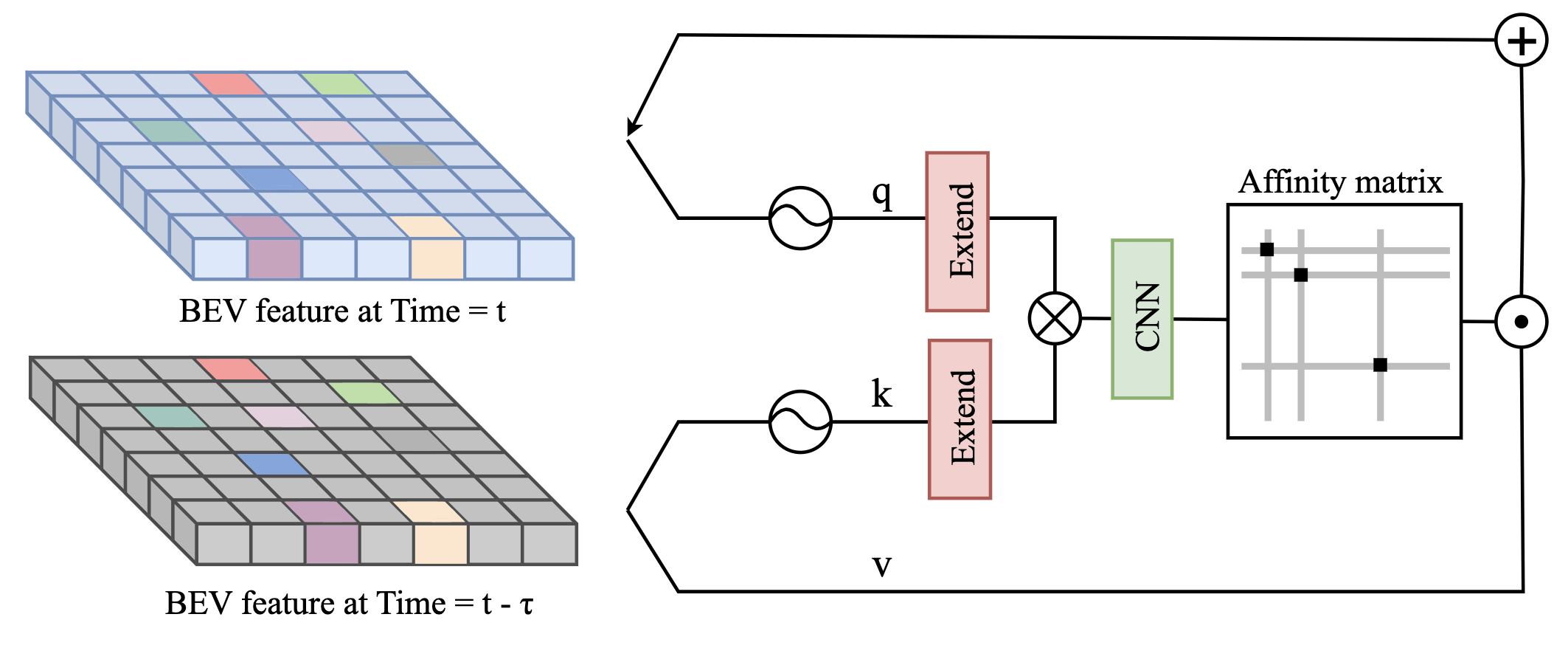} 
    \caption{Overview of the cross-attention module. It takes BEV features from two frames and extracts the features at pedestrian locations, which are used as tokens for the cross-attention processing. The module computes affinity scores between pedestrians in different frames and propagates the features across frames.  \raisebox{0.0000001\textheight}{\includegraphics[scale=0.026]{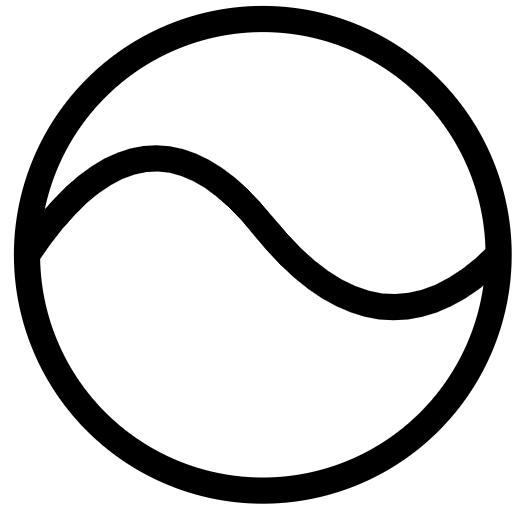}} represents positional encoding, while $\otimes$ and $\odot$ denote element-wise and matrix multiplication, respectively. }
    \label{fig:fig1}
\end{figure}

Moreover, prior methods \cite{ref36, ref38, ref39} have processed the BEV representation through a decoder based on Feature Pyramid Network (FPN) architecture. This approach expands the receptive field for each location, enabling the aggregation of appearance and location information from the distortion shadows. However, in traditional FPNs, the use of $1 \times 1$ convolution to fuse the upsampled and bottom-up feature maps creates a rigid structure, restricting its ability to adapt to geometric variations and deformations caused by perspective transformations. This negatively affects the accurate aggregation of location and appearance information in the BEV feature, leading to suboptimal predictions.

Transformers \cite{ref56} were introduced as a type of neural network that leverages attention mechanisms to model relationships between input elements. In the transformer architecture, attention is computed between query, key, and value tokens, where positional encoding is incorporated into the query and the key tokens to maintain the order of the tokens and address permutation invariance. This mechanism is referred to as self-attention when the query and the key originate from the same input, and cross-attention when they come from different inputs. In multi-object tracking, the objective is to establish associations between instances of the same object over multiple frames, which closely resembles the query-key concept in cross-attention mechanisms. The features of an object remain highly similar across different frames, enabling the query-key process to produce larger attention weights that serve as the association scores. In \cite{ref54}, this process is further enhanced by applying a CNN to the query-key pairs to more effectively capture complex patterns and interactions between each pair. Moreover, applying cross-attention between instances across frames enables the propagation of instances across frames \cite{ref57}. Upon examining the literature, we found that cross-attention techniques have been employed in single-view multi-object tracking \cite{ref54, ref57}, but they have not been explored in multi-view scenarios.

In this paper, we propose a novel approach that makes a step forward in tracking pedestrians from multiple views by leveraging the cross-attention mechanism \cite{ref56} within a multi-view tracking model. In particular, adopting an early-fusion strategy for detection, our method leverages a single cross-attention module to establish associations between pedestrians in different frames and propagate pedestrian features across frames, resulting in a more robust representation for each pedestrian. Figure \ref{fig:fig1} provides a visual overview of the cross-attention module. Moreover, by incorporating temporal information, our method goes beyond the limitations of the state-of-the-art multi-view tracker EarlyBird \cite{ref36}, which distinguishes pedestrians based on appearance features from a single frame, enabling more robust tracking across frames. To the best of our knowledge, this is the first time attention mechanisms are used in a multi-view pedestrian tracking setup. 

Furthermore, we propose a more robust decoder architecture to process the BEV feature, in which each layer downsamples the BEV feature by a factor of 2. The output of each layer is then upsampled to match the size of the previous layer. The features are concatenated and processed with deformable convolutions \cite{ref69}, which dynamically adjust the receptive field for each spatial location. This helps alleviate the spatial misalignment across scales in the fusion stage and enhances the aggregation of appearance and location information, resulting in a significant improvement in tracking performance. 

The key contributions of this paper are summarized as follows:

\begin{itemize}
    \item We propose adopting an attention mechanism for multi-view pedestrian tracking, a novel approach not explored in existing literature for tracking pedestrians from multiple views.
    \item We introduce a stronger decoder architecture for processing the BEV features, ensuring a more accurate and adaptive aggregation of location and appearance information.
    
    \item We evaluate our model's performance both quantitatively and qualitatively, demonstrating how it outperforms existing state-of-the-art methods, with an IDF1 score of $96.1\%$ on Wildtrack dataset, and $85.7\%$ on MultiviewX dataset.

\end{itemize}

\section{Related Work}
\label{sec:formatting}

\textbf{Single-view tracking} Owing to the significant advancements in object detection methods, tracking-by-detection has become the main approach for multi-object tracking, as it simplifies the process into two main tasks: (i) determining object locations in each frame separately, (ii) connecting the corresponding detections across different time steps to form tracks. SORT \cite{ref23} is a popular tracking-by-detection model that utilizes a Kalman filter for motion prediction and the Hungarian algorithm for association. DeepSORT \cite{ref24} was developed to overcome SORT limitations by augmenting its cost matrix with deep appearance features, integrating motion prediction and re-ID for better associations. ByteTrack \cite{ref25} introduces a flexible association strategy that associates every detected box, based on the observation that low-confidence detection boxes may indicate occluded objects. DAN \cite{ref27} computes affinities between objects in a pair of frames by analyzing all possible pairwise combinations of their appearance features using a CNN designed for affinity computations.

\begin{figure*}[t]
    \centering
    \includegraphics[width=0.99\textwidth]{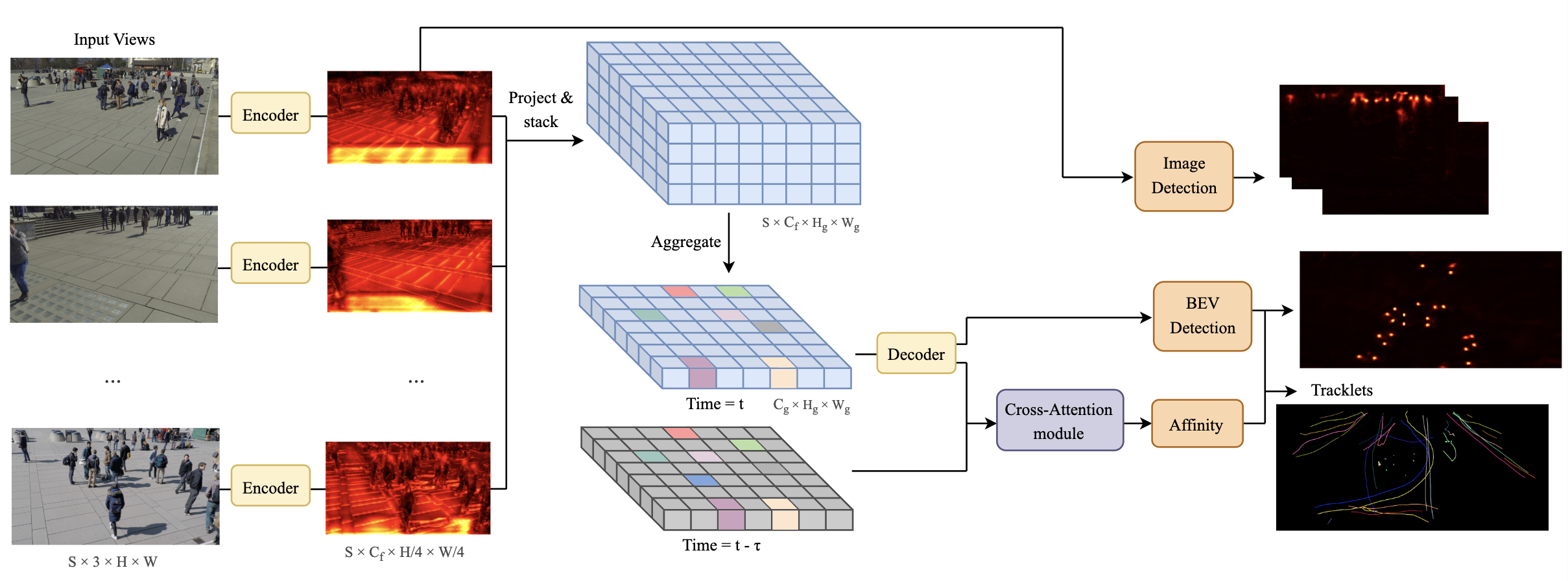} 
    \caption{Overview of our model pipeline. The feature maps are extracted from the input views, projected onto a common ground plane, and aggregated to form a unified BEV feature, which is then passed into the decoder for further processing. The BEV features from two neighboring frames are then processed by the cross-attention module to efficiently establish associations and propagate information across frames.}
    \label{fig:fig2}
\end{figure*}

One-shot trackers form a specific category of multi-object tracking, formulating it as a multi-task learning framework by optimizing detection and tracking simultaneously within a single network, leading to shorter inference time. However, their performance is often inferior compared to two-step trackers. The features predicted for tracking can be categorized into two types: appearance features \cite{ref12, ref14, ref16, ref54} or motion features \cite{ref19, ref6, ref22}. In the context of appearance-based tracking, TrackR-CNN \cite{ref12} extends Mask R-CNN \cite{ref13} by incorporating a re-ID feature head, along with bounding box regression and mask generation heads for each proposal. Likewise, JDE \cite{ref14} extends YOLOv3 \cite{ref15} detector by adding a re-ID feature head. FairMOT \cite{ref16} builds upon CenterNet \cite{ref17} detector by integrating a re-ID head along with the bounding box center, offset, and size prediction heads. Notably, FairMOT stands out from other methods due to its anchor-free design that relies on detecting objects using a single detection point rather than bounding boxes, which leads to enhanced separation of re-ID features. TicrossNet \cite{ref54} uses a single cross-attention module to complete all key tracking processes to enable real-time tracking. Shifting to motion-based tracking, D\&T \cite{ref19} extends the R-FCN framework \cite{ref20} by incorporating an RoI-tracking layer that estimates the inter-frame offsets of the bounding boxes. Tracktor \cite{ref6} leverages the detector's regression head to achieve temporal alignment of bounding boxes, which enables the consistent propagation of object identities across frames. CenterTrack \cite{ref22} builds upon CenterNet \cite{ref17} detector by predicting the offsets of object centers between adjacent frames utilizing a triplet input comprising the current and previous frames, along with the heatmap of detections from the previous frame. However, motion-based trackers primarily focus on associating objects in neighboring frames and do not reinitialize lost tracks, thereby hindering their ability in handling occlusions.

\medskip

\noindent\textbf{Multi-view tracking} 
Most multi-target multi-camera (MTMC) trackers operate under the assumption that the cameras are synchronized and have overlapping fields of view. Fleuret \textit{et al.} \cite{ref1} first compute the probabilistic occupancy map (POM) and merge it with motion and color attributes to track objects using the Viterbi algorithm. In \cite{ref2}, data association is reformulated as a flow optimization task, solving it with the K-Shortest Paths (KSP) algorithm. On the other hand, several graph-based approaches have been proposed to model correspondences across views for the MTMC problem.  In \cite{ref3}, hypergraphs were leveraged to model the correspondences between the camera views, and subsequently solved using a min-cost algorithm. Similarly, \cite{ref4, ref5} used a multi-commodity network flow formulation, with the tracking problem solved using the branch-and-price algorithm \cite{ref4} or the min-cost algorithm \cite{ref5}

Lately, a two-step approach has become widely adopted. The first step involves producing local tracklets for objects from each camera separately, a process commonly referred to as single-camera MOT. For the second step, a range of techniques have been developed to associate local tracklets across different cameras. Some approaches \cite{ref7, ref8} exploit epipolar geometry principles to establish correspondences across the views. In \cite{ref7}, tracking of people's head centers is proposed to handle occlusion in dense crowds, while \cite{ref8} matches people across views using their principal axes. On the other hand, \cite{ref9} combined appearance and motion cues with ground plane locations, reformulating the association step as a hierarchical composition optimization problem. While the two-step approach offers benefits such as a reduced hypothesis space, they are prone to ID-switch errors in local tracklets, where distinct objects can be incorrectly combined into the same trajectory. To overcome these challenges, many models \cite{ref10, ref11} have rearranged the first two steps by first projecting 2D detections onto the ground plane and create a graph structure with nodes containing re-ID features. Associations between the nodes are formed using graph neural networks for link prediction. In \cite{ref10}, spatial and temporal graphs are optimized separately, whereas in \cite{ref11}, the optimization occurs simultaneously across both domains.

Recent state-of-the-art models \cite{ref36, ref38, ref39} reflect the concept of one-shot trackers and distinguish themselves from earlier methods by utilizing an early-fusion strategy \cite{ref28} for detection.  A notable advantage of early-fusion approaches is that they can be trained in an end-to-end manner. Moreover, they significantly improve detection quality compared to late-fusion methods \cite{ref36}. One such example is EarlyBird \cite{ref36}, which extends FairMOT's \cite{ref16} joint detection and re-ID extraction to multi-view tracking, performing detection and re-ID feature extraction in the BEV space. On the other hand, MVFlow \cite{ref38} integrates human flow constraints into a deep learning framework by developing a weakly supervised approach to predict human flow over time using only detection data on the ground plane. In addition, TrackTacular \cite{ref39} explores the use of three lifting strategies: Simple-BEV \cite{ref40}, BEVFormer \cite{ref41}, and Lift-Splat-Short \cite{ref42}, to lift the camera features to a common BEV space. For tracking, the method utilizes the historical BEV feature to estimate the position of each detection in the prior frame, effectively learning the associations between detections at consecutive time steps. 

Building on the successes of early-fusion strategies and leveraging the power of attention mechanisms, our proposed model addresses the lack of robustness in appearance features extracted from the BEV space by employing a cross-attention module to establish robust associations between pedestrians across frames, while effectively propagating pedestrian features across frames to ensure a more robust feature representation for each pedestrian.

\section{Methodology}

The overall structure of the proposed model is shown in Figure \ref{fig:fig2}. Given a set of synchronized images from $S$ calibrated cameras, these images are augmented and passed into the encoder to extract the feature maps. Next, the feature maps are projected to a common BEV space using perspective projection, followed by aggregating the feature maps from all views to yield a unified BEV feature, which is then fed to the decoder. Furthermore, the key contribution of our approach is the cross-attention module, which takes the BEV features from two neighbouring frames to establish associations between pedestrian instances and propagate pedestrian features across frames.

\subsection{Encoder}
The encoder, as done in \cite{ref36}, extracts feature maps from input images provided by a set of $S$ cameras using ResNet, each with a size of $3 \times H \times W$. It includes three blocks of ResNet that progressively downsample the images by a factor of 2. The features from each layer are upsampled and combined with the outputs from the previous layer, then passed through a convolutional layer. This results in the final feature with dimensions of $C_{f} \times H_{f} \times W_{f}$, where $H_{f} = H/4$, $W_{f} = W/4$, and $C_{f}$ denotes the output feature channels.

\subsection{Projection}
The feature maps from all $S$ cameras are transformed into a unified space using perspective projection, creating a 2D BEV feature map of the scene. Using the pinhole camera model \cite{ref55}, the mapping from 3D points $(x, y, z)$ to 2D pixel coordinates $(u, v)$ in the image plane is performed using a $3 \times 4$ perspective transformation matrix $P = K \left[ R \mid t \right]$, as follows:
\begin{equation}
\scalebox{0.9}{$
d \begin{pmatrix} u \\ v \\ 1 \end{pmatrix} = K[R|t] \begin{pmatrix} x \\ y \\ z \\ 1 \end{pmatrix} = \begin{bmatrix} p_{11} & p_{12} & p_{13} & p_{14} \\ p_{21} & p_{22} & p_{23} & p_{24} \\ p_{31} & p_{32} & p_{33} & p_{34} \end{bmatrix} \begin{pmatrix} x \\ y \\ z \\ 1 \end{pmatrix}$}
\label{eq:eq1},
\end{equation}
where $d$ represents a scaling factor accounting for the feature maps downsampling, while $[R | t]$ refers to the extrinsic parameter matrix and $K$ denotes the intrinsic parameter matrix. In our method, the features are mapped onto the ground plane at $z=0$, as done in \cite{ref28}, simplifying the projection to:
\begin{equation}
\scalebox{0.9}{$
d \begin{pmatrix} u \\ v \\ 1 \end{pmatrix} = P^{(s)}_{0} \begin{pmatrix} x \\ y \\ 1 \end{pmatrix}  =  \begin{bmatrix}
 p_{11} &  p_{12} &  p_{14} \\
 p_{21} &  p_{22} &  p_{24} \\
 p_{31} &  p_{32} &  p_{34} 
\end{bmatrix} \begin{pmatrix} x \\ y \\ 1 \end{pmatrix}$}
\label{eq:eq2},
\end{equation}
 where $P^{(s)}_{0} = K [R | t]$ is a $3 \times 3$ transformation matrix for camera $s$, excluding the third column of $P$. All feature maps are projected onto a common ground plane, which is discretized into a grid of predetermined dimensions $[H_g, W_g]$, based on the size of the annotated area. Each position in the grid corresponds to a $10 \, \text{cm} \times 10 \, \text{cm}$ area in the real world, which is further downsampled by a factor of 4 to ensure efficient memory usage. Lastly, a convolutional layer is applied to aggregate the features from all views into a single feature representation of size 
$C_{g} \times H_{g} \times W_{g}$.

\subsection{Decoder}

In the decoder architecture, we build upon the work in \cite{ref36, ref39} by utilizing a ResNet-18 decoder. In this framework, each layer of ResNet downsamples the BEV feature by a factor of 2. The output from each layer is then upsampled to match the size of the preceding layer output. The two features are concatenated and passed through a deformable convolution layer \cite{ref69}. Unlike conventional FPN structures which use $1 \times 1$ convolutions for feature fusion, we use $3 \times 3$ deformable convolutions, which offer greater flexibility by adaptively adjusting the receptive field for each location, thereby improving the multi-scale feature alignment in the fusion stage and leading to more precise aggregation of location and appearance features of pedestrians. The output from the feature pyramid network has the same dimensions as the input $C_{g} \times H_{g} \times W_{g}$, but expands the receptive field for each spatial location.

\subsection{Cross-Attention Module}
The cross-attention module is the central component introduced in our model for tracking pedestrians from multiple views. This component facilitates the robust association of pedestrians across frames and enables effective feature propagation, ensuring more robust feature representation.
The decoded BEV features from the current and neighboring frames, $X_{t}$ and $X_{t-\tau}$ (with $1 \leq \tau \leq 3 $ in our experiments), are fed into the cross-attention module, which is illustrated in Figure \ref{fig:fig1}. We apply 3D positional encoding \cite{ref58}, to enrich the BEV feature with spatial and temporal information, resulting in positionally encoded BEV features $Pe(X_{t})$ and $Pe(X_{t-\tau})$. The pedestrian features at the center of each detection in the BEV feature are extracted and used as tokens. The tokens from the current frame $Pe(X_{t})$ serve as the queries ${Q} \in \mathbb{R}^{N_t \times C_f}$, while the tokens from the previous frame $Pe(X_{t-\tau})$ represent the keys ${K} \in \mathbb{R}^{N_{t-\tau} \times C_f}$. The tokens from the original representation $X_{t-\tau}$ are used as values ${V} \in \mathbb{R}^{N_{t-\tau} \times C_f}$, with $N_{t}$ and $N_{t-\tau}$ representing the number of tokens in the current and previous frames, respectively. Inspired by \cite{ref54}, the similarities between the queries and keys are efficiently encoded using a CNN. To achieve this, the queries and keys are expanded by duplicating their elements to create token-level pairs, resulting in ${Q} \in \mathbb{R}^{N_t \times N_{t-\tau} \times C_f}$ and ${K} \in \mathbb{R}^{N_t \times N_{t-\tau} \times C_f}$. These are then combined through element-wise multiplication, represented as:
\begin{equation}
M = Q \otimes K, \quad  {M} \in \mathbb{R}^{N_t \times N_{t-\tau} \times C_f},
\end{equation} where $\otimes$ denotes element-wise multiplication.  $M$ is passed through a CNN that encodes the affinity between each query-key pair. The architecture of the CNN, which consists of three $1 \times 1$ convolutional layers with batch normalization and ReLU activation, is detailed in Table~\ref{tab:micro_cnn_specs}. This CNN reduces the dimensionality of $M$ to a single-channel representation, which captures the similarity scores between query-key pairs. The resulting matrix ${A} \in \mathbb{R}^{N_t \times N_{t-\tau}}$ encodes the affinity between pedestrians across different frames. However, it does not take into account the pedestrians entering or leaving the video between the two frames. To handle these cases, we extend $A$ by appending an extra row and an extra column \cite{ref27}, with values $v = \gamma\mathbf{1}$, where $\gamma$ is a learnable parameter optimized during training and $\mathbf{1}$ is a vector of ones. This results in two matrices ${A_1} \in \mathbb{R}^{(N_t+1) \times N_{t-\tau}}$ and ${A_2} \in \mathbb{R}^{N_t \times (N_{t-\tau}+1)}$, respectively. Consequently, column-wise softmax \cite{ref27} is applied to ${A_1}$, yielding $\hat{A_1}$, which represents associations for each pedestrian in frame $X_{t-\tau}$ to all $N_t+1$ identities. Similarly, row-wise softmax \cite{ref27} is applied to $A_2$, producing $\hat{A_2}$, which encodes associations between each pedestrian in frame $X_{t}$ and all $N_{t-\tau}+1$ identities.  

\begin{table}[t]
    \centering
    \renewcommand{\arraystretch}{1} % Adjust row height as needed
    \begin{tabular}{@{}cccc@{}}
        \toprule
        Layer & I.C. & O.C. & Kernel Size \\ 
        \midrule
        1 & 128 & 64 & 1×1 \\
        2 & 64 & 32 & 1×1 \\
        3 & 32 & 1 & 1×1 \\
        \bottomrule
    \end{tabular}
    \captionsetup{justification=raggedright}
    \caption{Specifications of the CNN, where I.C. and O.C. represent the input and output channels, respectively.}
    \label{tab:micro_cnn_specs}
\end{table}

We aim to predict a single affinity matrix, but $\hat{A_1}$ and $\hat{A_2}$ have incompatible dimensions and cannot be combined by element-wise multiplication. Therefore, we multiply the inner matrices as $A_{\text{aff}} = \hat{A_1} \otimes \hat{A_2}$, where $\hat{A_1},\hat{A_2} \in \mathbb{R}^{N_t \times N_{t-\tau}}$. The extra row and column are handled separately by multiplying them with the mean of their respective rows and columns in matrix $A_{\text{aff}}$, to ensure that they match its distribution. It is important to note that although TicrossNet \cite{ref54} combined the matrices resulting from softmax operations through element-wise multiplication, their approach does not handle the objects entering or leaving the video between the two frames. The final affinity matrix, $A_{\text{aff}} \in \mathbb{R}^{(N_t+1) \times (N_{t-\tau}+1)}$ is obtained by appending the adjusted extra row and column. Next, the cross-attention mechanism uses the inner affinity matrix, $A_{\text{aff}} \in \mathbb{R}^{N_t \times N_{t-\tau}}$, to guide the propagation of instances across frames. It is multiplied with the value tokens $V$ as follows:
\begin{equation}
\hat{V} = A_{\text{aff}} \odot V, \quad \hat{V} \in \mathbb{R}^{N_{t} \times C_f},
\end{equation}
where $\odot$ denotes matrix multiplication. The result from the attention, $\hat{V}$, contains the attended tokens, capturing the relevant features of the previous frame for each query in the current frame. By multiplying the affinity matrix with the value tokens, each query (pedestrian's feature in the current frame) attends to the most similar key-value pair (pedestrian's feature in the previous frame). The resulting attended tokens from the attention are then added to the original positions of query tokens in the current frame $X_t$. This yields the refined BEV feature, $\hat{X_t}$, which enhances each pedestrian's feature by incorporating relevant information from the previous frames. Moreover, this makes our approach stand out from the state-of-the-art tracker, EarlyBird \cite{ref36}, which relies on the appearance feature from a single frame to differentiate between pedestrians. The refined BEV feature serves as input to the attention mechanism for the subsequent frames and provides the CNN with more powerful and discriminative features, enabling it to better estimate the affinity between each query-key pair.

\subsection{Heads and losses}
We adopt the CenterNet \cite{ref80} architecture to obtain the POM on the ground plane. This is achieved by applying a center prediction head to the BEV feature map, which allows the model to predict the center locations of pedestrians, resulting in an output size of $1 \times H_g \times W_g$. An offset prediction head is also used to predict an offset map to compensate for the loss of decimal part during downsampling. This results in an offset map of size $2 \times H_g \times W_g$. We optimize the center head using Focal Loss \cite{ref59}, while the offset head is optimized uing L1 loss. Each head is composed of a $3 \times 3$ convolution, an activation layer, and a $1 \times 1$ convolution to yield the final size.

We also incorporate detection heads for the image features to provide additional supervision by predicting the 2D bounding box centers and the approximated foot location at the bottom midpoint of the bounding box. For tracking, the estimated affinity matrix is optimized using cross-entropy loss. Moreover, we use uncertainty loss to balance the detection and tracking losses before summing them up, following FairMOT \cite{ref16}.

\subsection{Inference}
The synchronized images are processed through the model pipeline to obtain the refined BEV feature. The refined BEV features is passed to the center head to obtain the POM. Non-maximum suppression is then applied using $3 \times 3$ max pooling operation \cite{ref22}, and only detections with a confidence score above a threshold of $\theta=0.5$ are extracted. The BEV features at each detection location are extracted from the BEV feature map and passed as tokens to the cross-attention module and the similarity matrix $A_{\text{aff}}$ is computed. 

We employ the hierarchical online data association method presented in MOTDT \cite{ref60}, focusing on tracking the centers of pedestrian in the BEV. To begin, we initialize a series of tracklets using the centers detected in the first timestep. For each next timestep, the detected centers are linked with the tracklets using a matching strategy consisting of two stages.

The first stage involves using the Kalman Filter \cite{ref61} and the affinity matrix $A_{\text{aff}}$, excluding the extra row and column, to produce the initial tracking results. In particular, we employ the Kalman Filter to estimate tracklet positions in the next frame and calculate the Mahalanobis distance ($D_{m}$) between the estimated and detected centers, as described in DeepSORT \cite{ref24}. We transform the affinity matrix into a distance matrix ($D_{A}$) by subtracting the maximum value from all values in the matrix. Then, the two distance matrices are combined using a weighted sum, $D = \lambda D_{A} + (1 - \lambda) D_{m}$, where the hyperparameter $\lambda$ controls the contribution of each metric, and is set to 0.98 in our experiments. Following the JDE protocol \cite{ref14}, if the Mahalanobis distance exceeds a specific threshold, it is set to infinity, which prevents tracking of trajectories with unrealistic motion. Lastly, the Hungarian algorithm is used to find the optimal matchings from the cost matrix $D$, using a matching threshold of $\tau_1 = 0.5$.

At the second stage, the unmatched detections are linked with unmatched tracklets based on the distances between their centers, with a higher matching threshold of $\tau = 1.8\,\text{m}$. If no association is made, unmatched detections are established as new tracklets, while unmatched tracklets are retained for 10 frames to allow for re-association if they reappear later.

\begin{table*}[t]
    \centering
    \renewcommand{\arraystretch}{1.15} % Adjust row height as needed
    \begin{tabular}{@{}l@{\hskip 5em}cccccc@{}} % Increase hskip to 1em
        \toprule
        & \multicolumn{5}{c}{Wildtrack} \\ 
        \cmidrule(lr){2-6}
        & IDF1↑ & MOTA↑ & MOTP↑ & MT↑ & ML↓ \\ 
        \midrule
        KSP-DO  \cite{ref49} & 73.2 & 69.6 & 61.5 & 28.7 & 25.1 \\
        KSP-DO-ptrack   \cite{ref49} & 78.4 & 72.2 & 60.3 & 42.1 & \underline{14.6} \\
        GLMB-YOLOv3   \cite{ref63}   & 74.3 & 69.7 & 73.2 & 79.5 & 21.6 \\
        GLMB-DO   \cite{ref63}   & 72.5 & 70.1 & 63.1 & \textbf{93.6} & 22.8 \\
        DMCT       \cite{ref64}      & 77.8 & 72.8 & 79.1 & 61.0 & \textbf{4.9} \\
        DMCT Stack    \cite{ref64}   & 81.9 & 74.6 & 78.9 & 65.9 & \textbf{4.9} \\ 
        $\text{ReST}^{\dagger}$ \cite{ref10}  & 86.7 & 84.9 & 84.1 & \underline{87.8} & \textbf{4.9} \\
        EarlyBird \cite{ref36}      & 92.3 & 89.5 & \underline{86.6} & 78.0 & \textbf{4.9} \\
        MVFlow   \cite{ref38}       & 93.5 & 91.3 & {-} & {-} & {-} \\
        TrackTacular  \cite{ref39}  & \underline{95.3} & \underline{91.8} & 85.4 & \underline{87.8} & \textbf{4.9} \\
        \midrule
        Ours    & \textbf{96.1} & \textbf{92.7} & \textbf{88.8} & \underline{87.8} & \textbf{4.9} \\ 
        \hline\hline
       \\[-7pt] 
        & \multicolumn{5}{c}{MultiviewX} \\ 
        \cmidrule(lr){2-6}
        & IDF1↑ & MOTA↑ & MOTP↑ & MT↑ & ML↓ \\ 
        \midrule
        EarlyBird  \cite{ref36} &  82.4 & 88.4  & \underline{86.2} & 82.9 &  \textbf{1.3} \\
        TrackTacular \cite{ref39} & \underline{85.6}  & \textbf{92.4} & 80.1 & \textbf{92.1} & \underline{2.6} \\
        \midrule
        Ours    & \textbf{85.7} & \underline{91.3} & \textbf{86.9} & \underline{89.5} & \textbf{1.3} \\  
        \bottomrule
    \end{tabular}
    \captionsetup{justification=raggedright}
    \caption{Tracking performance evaluation on Wildtrack and MultiviewX datasets against state-of-the-art models, with the best and second-best performers in bold and underlined, respectively. The results for MultiviewX dataset are available only for \cite{ref36} and \cite{ref39}. $ ^{\dagger}$ ReST results are re-computed by \cite{ref36}.}
    \label{tab:performance_metrics2}
\end{table*}

\section{Experiments}
\subsection{Datasets}

\noindent \textbf{Wildtrack Dataset} Wildtrack \cite{ref49} is a dataset recorded in a real-world environment with seven fixed cameras, each recording at $1080 \times 1920$ resolution and $2$ fps. The cameras are synchronized and calibrated with overlapping fields of view that cover a region of $12\,\text{m} \times 36\,\text{m}$. The region is divided into a $480 \times 1440$ grid, with each grid cell measuring $2.5\, \text{cm} \times 2.5\, \text{cm}$. On average, each frame contains 20 pedestrians, with each location being captured by approximately $3.74$ cameras. The videos capture a public environment with unscripted pedestrian movement. The dataset is contains $400$ frames, with the first $360$ allocated for training the model and the remaining $40$ for testing.

\medskip
\noindent \textbf{MultiviewX Dataset} MultiviewX \cite{ref28} is a synthetic dataset constructed using a game engine. It is designed to resemble Wildtrack in terms of resolution, frame rate, and frame count, but with one fewer camera. The area covered is $16\,\text{m} \times 25\,\text{m}$, divided into a $640 \times 1000$ grid, maintaining the same cell size. Each frame captures about 40 pedestrians, roughly twice the number in Wildtrack, and each location is monitored by around $4.41$ cameras. Similar to Wildtrack, this dataset consists of $400$ frames, with the first $360$ used for training and the last $40$ for testing.

\subsection{Evaluation Metrics}
Tracking metrics are calculated on the detected center points on the ground plane. We report CLEAR MOT metrics \cite{ref50} and identity-aware metrics \cite{ref51}. In particular, we consider IDF1, Multiple Object Tracking Accuracy (MOTA), and Multiple Object Tracking Precision (MOPT). Moreover, we report Mostly Tracked (MT) and Mostly Lost (ML) as percentages of the total number of pedestrians in the test set. For a positive assignment, the threshold is set to $r = 1\, \text{m}$. Following \cite{ref36, ref39}, we emphasize IDF1 and MOTA as the main performance metrics. MOTA considers missed detections, false detections, and identity switches. IDF1 focuses on trajectory-level identity matching and accounts for false positives, missed detections, as well as identity switches.

\begin{table}[t]
    \centering
    \renewcommand{\arraystretch}{1.2} % Adjust row height as needed
    \begin{tabular}{@{}lccc@{}}
        \toprule
        Configuration & IDF1 & MOTA & MOTP \\ 
        \midrule
        Baseline & 94.2 & 90.4 & \underline{89.0} \\
        + Cross-Attention Module & \underline{95.9} & \underline{92.1} & \textbf{89.2} \\
        + Optimized Decoder & \textbf{96.1} & \textbf{92.7} & 88.8 \\
        \bottomrule
    \end{tabular}
    \captionsetup{justification=raggedright}
    \caption{Evaluation of different components of our model on Wildtrack dataset.}
    \label{tab:ablation_study2}
\end{table}

\subsection{Implementation Details}

The input images have a resolution of $720 \times 1280$ pixels.  We apply a set of augmentations to the images during training to avoid overfitting and enhance generalization. Each image undergoes random resizing within a scale range of $[0.8,1.2]$, is shifted by a random offset from the center, and cropped to maintain its original dimensions \cite{ref29, ref40}. The camera intrinsics matrices $K$ are adjusted accordingly to maintain multi-view consistency. In order to avoid overfitting the detector, some noise is introduced to the translation vector $t$ of the extrinsic matrix, following the approach in \cite{ref36, ref39}. In the training setup, we use Adam optimizer and incorporate a one-cycle learning rate scheduler, setting the highest learning rate to $1 \times 10^{-3}$. The encoder and decoder networks are initialized using weights pre-trained on \textit{ImageNet-1K}.

\begin{figure*}[t]
  \centering
  % First figure
  \begin{subfigure}{0.49\textwidth}
    \centering
    \includegraphics[width=\textwidth]{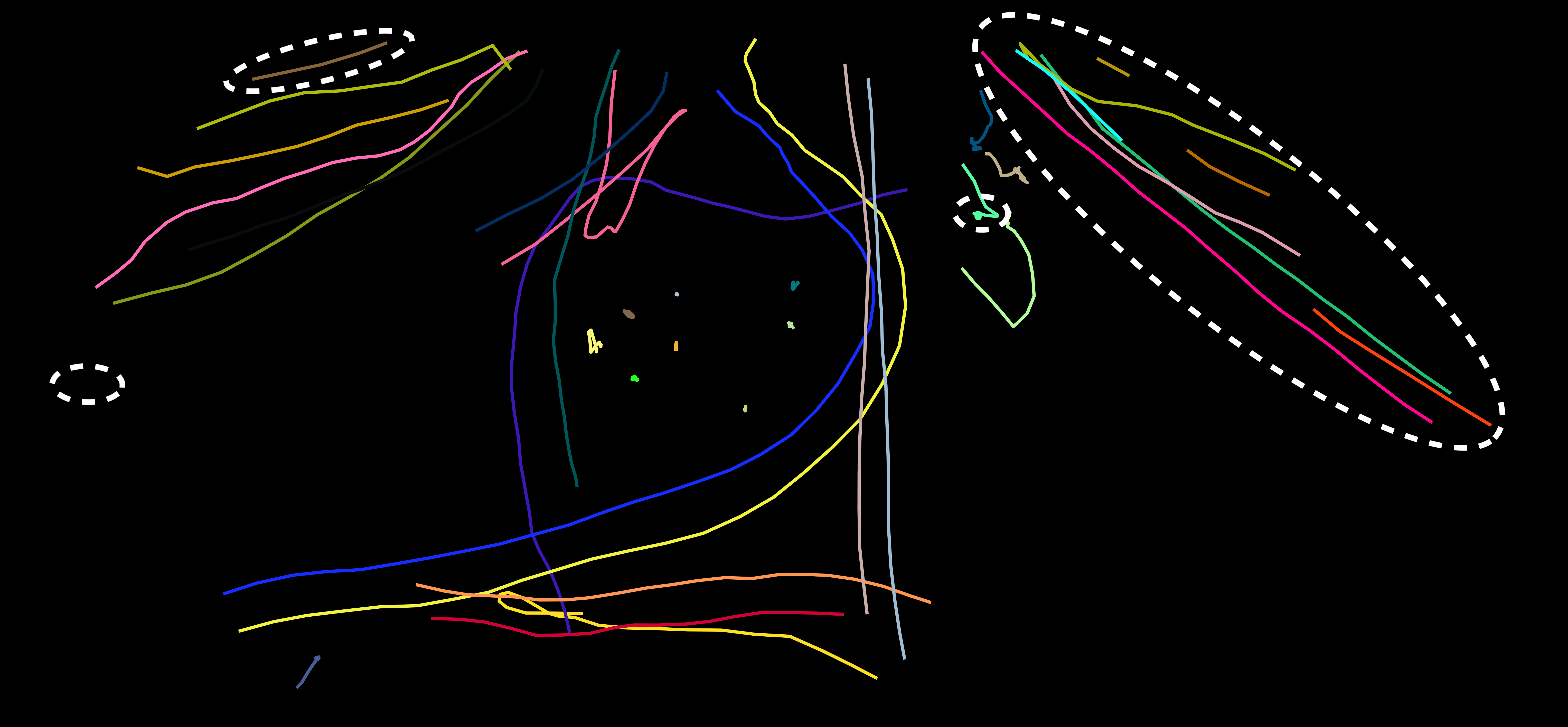}
    \caption{Our model trajectories.}
    \label{fig:prediction}
  \end{subfigure} \hfill
  % Second figure
  \begin{subfigure}{0.49\textwidth}
    \centering
    \includegraphics[width=\textwidth]{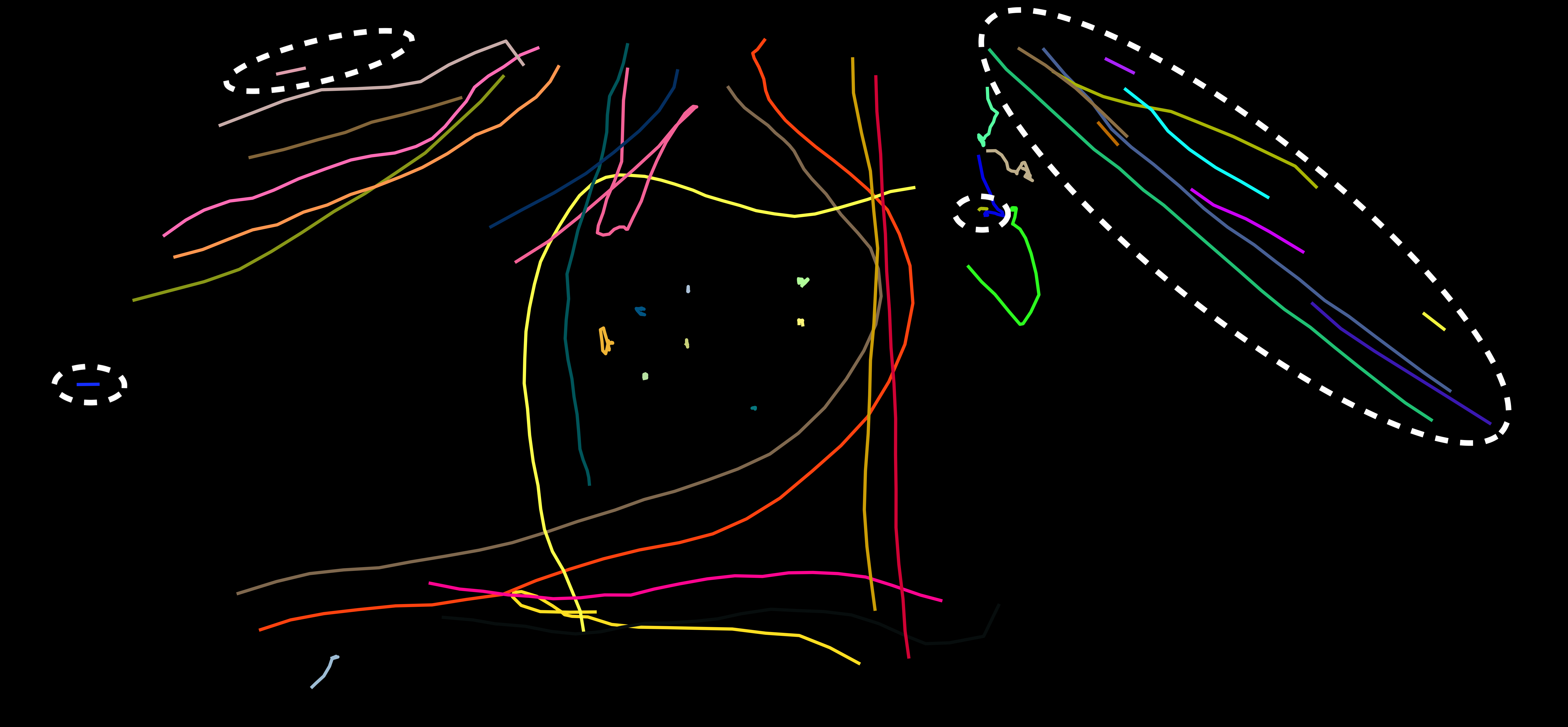}
    \caption{Baseline model trajectories.}
    \label{fig:ground_truth}
  \end{subfigure} \hfill
  % Third figure
    \vspace{2mm} 
  \begin{subfigure}{0.49\textwidth}
    \centering
    \includegraphics[width=\textwidth]{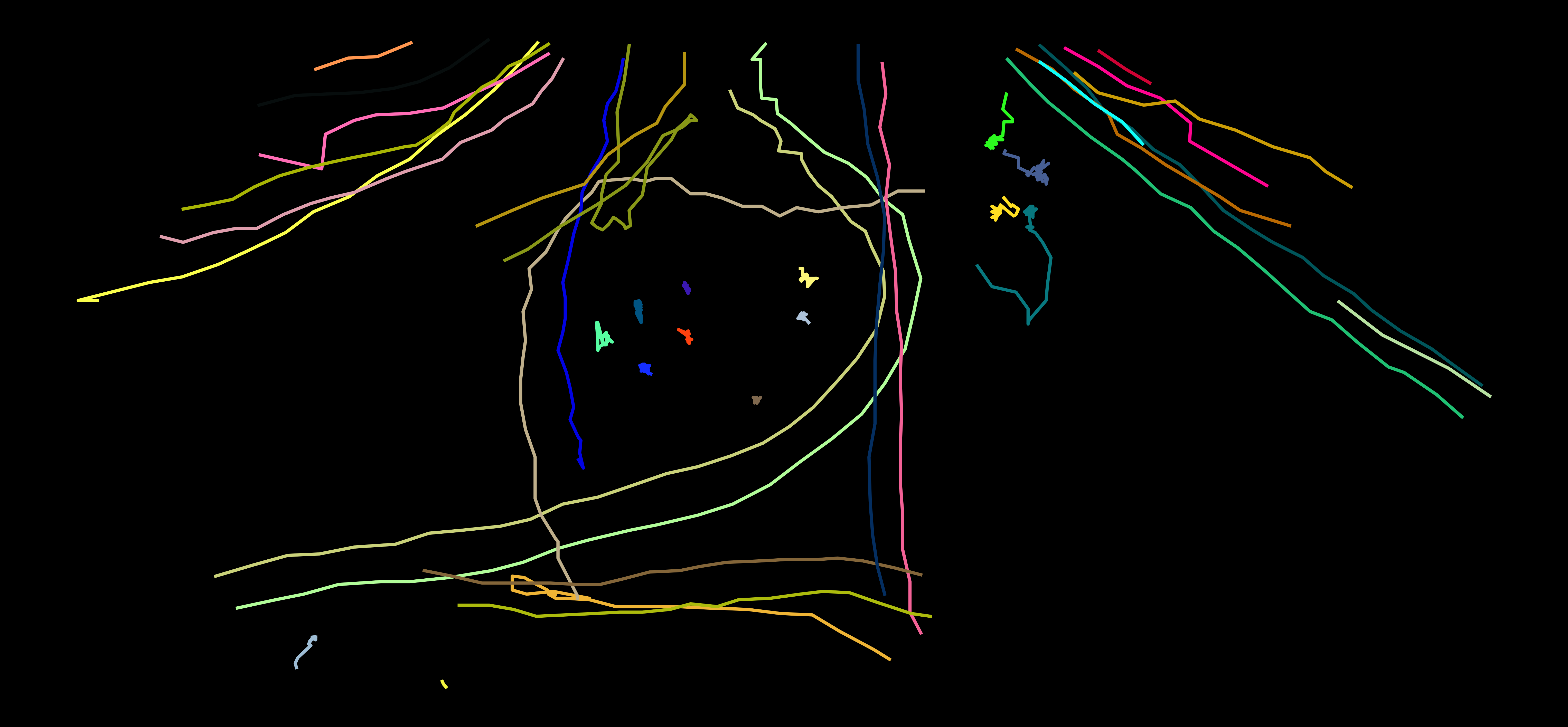}
    \caption{Ground truth trajectories.}
    \label{fig:third_image}
  \end{subfigure}
  
  \caption{Comparison of tracking performance between our approach and the baseline using the full Wildtrack test set. The dashed ovals highlight the areas where our model demonstrates improved tracking accuracy compared to the baseline model.}
  \label{fig:tracking_vis}
\end{figure*}

\subsection{Quantitative Results}

The evaluation of our model against state-of-the-art models on Wildtrack and MultiviewX datasets is shown in Table~\ref{tab:performance_metrics2}. Our model achieves state-of-the-art results on both datasets, demonstrating significant improvements in comparison to the three latest approaches that perform tracking in the BEV space, namely, EarlyBird \cite{ref36}, MVFlow\cite{ref38}, and TrackTacular \cite{ref39}. On Wildtrack dataset, we improve IDF1 and MOTA by $+0.8\%$ and $+0.9\%$, respectively. On MultiviewX dataset, our model shows an improvement in IDF1 score, with TrackTacular \cite{ref39} being the only method that achieves a higher MOTA score. In addition, our model improves MOTP in both dataset, with $+2.2\%$ increase in Wildtrack and $+0.7\%$ in MultiviewX.

\subsection{Ablation Studies}
We evaluate the impact of each component proposed in our model in Table~\ref{tab:ablation_study2}. The first row shows the performance of the baseline model, which is trained for detection only and Kalman filter is used for linking the detections. The second row presents the performance of our model when incorporating the cross-attention module. This addition results in a significant improvement in tracking performance, already advancing the model’s performance beyond state-of-the-art models. Furthermore, the last row shows the impact of utilizing the optimized decoder, which incorporates deformable convolution in the fusion stage of the FPN structure. The optimized decoder primarily boosts detection performance, leading to an improvement in MOTA from $92.1\%$ to $92.7\%$, as MOTA is more sensitive to detection accuracy.

\subsection{Qualitative Analysis}

Figure \ref{fig:tracking_vis} provides a qualitative comparison between the performance of our proposed model and the baseline on the full Wildtrack test set. The figure illustrates the predicted trajectories, each corresponding to distinct pedestrian identities, plotted on the ground plane alongside the ground truth, following \cite{ref36}. We contrast the performance of our model, which integrates the cross-attention module, with the baseline model that relies on the Kalman filter for linking detections across frames. Notably, regions highlighted with dashed ovals emphasize specific instances where our model demonstrates superior performance, showcasing its ability to maintain better tracking consistency.

\section{Conclusion}

We present a robust multi-view tracking model that leverages the latest advancements in multi-view detection and incorporates cross-attention mechanisms to enhance tracking robustness. The cross-attention mechanism establishes associations between instances across frames and efficiently propagates features, resulting in a more robust representation of each pedestrian. Additionally, our optimized decoder architecture enhances the processing of BEV features, enabling more adaptive feature aggregation from object deformations and improving the accuracy of tracking. The experimental results show performance improvements over existing state-of-the-art models.

Among different multi-view tracking methods, tracking in the BEV space currently achieves the highest performance. However, it requires accurate 3D annotations and camera calibrations. Therefore, we could not evalute our model on older benchmark datasets such as CAMPUS \cite{ref9} and PETS09 \cite{ref81}. As a result, future work may explore methods to eliminate the dependence on strict calibration and annotation requirements.

{
    \small
    \bibliographystyle{ieeenat_fullname}
    \bibliography{main}
}

\end{document}